\documentclass{article} 

\usepackage{url}
\usepackage{bbm}

\usepackage{amssymb}
\usepackage{bbm}
\usepackage{amsthm}
\usepackage{amsfonts}       
\usepackage{nicefrac}       
\usepackage{microtype}      
\usepackage{graphicx}
\usepackage{caption}
\usepackage{subcaption}

\usepackage{amsmath}
\usepackage{wrapfig}
\usepackage{bigints}
\usepackage{hyperref}
\usepackage{booktabs}       
\usepackage{enumitem}

\usepackage{multirow}
\usepackage[export]{adjustbox}
\usepackage{titlesec}

\usepackage{amsmath,amsfonts,bm}

\newcommand{\figleft}{{\em (Left)}}

\newcommand{\figright}{{\em (Right)}}

\def\eqref#1{equation~\ref{#1}}

\def\1{\bm{1}}

\DeclareMathAlphabet{\mathsfit}{\encodingdefault}{\sfdefault}{m}{sl}
\SetMathAlphabet{\mathsfit}{bold}{\encodingdefault}{\sfdefault}{bx}{n}

\def\gD{{\mathcal{D}}}

\def\gF{{\mathcal{F}}}

\def\gH{{\mathcal{H}}}

\def\gS{{\mathcal{S}}}

\newcommand{\E}{\mathbb{E}}

\DeclareMathOperator*{\argmax}{arg\,max}
\DeclareMathOperator*{\argmin}{arg\,min}

\newtheorem{theorem}{Theorem}
\newtheorem{lemma}[theorem]{Lemma}
\theoremstyle{definition}
\newtheorem{definition}{Definition}

\usepackage[accepted]{icml2020}

\icmltitlerunning{Unsupervised Meta-Learning for Reinforcement Learning}

\begin{document}

\twocolumn[
\icmltitle{Unsupervised Meta-Learning for Reinforcement Learning}

\icmlsetsymbol{equal}{*}

\begin{icmlauthorlist}
\icmlauthor{Abhishek Gupta}{equal,goo}
\icmlauthor{Benjamin Eysenbach}{equal,ed}
\icmlauthor{Chelsea Finn}{to}
\icmlauthor{Sergey Levine}{goo}
\end{icmlauthorlist}

\icmlaffiliation{to}{Stanford University}
\icmlaffiliation{goo}{UC Berkeley}
\icmlaffiliation{ed}{Carnegie Mellon University}

\icmlcorrespondingauthor{Abhishek Gupta}{abhigupta@eecs.berkeley.edu}

\icmlkeywords{Reinforcement Learning, Meta Learning}

\vskip 0.3in
]




\printAffiliationsAndNotice{\icmlEqualContribution}

\begin{abstract}
Meta reinforcement learning (meta-RL) algorithms leverage experience from learning previous tasks to learn how to learn new tasks quickly. However, this process requires a large number of meta-training tasks to be provided for meta-learning. In effect, meta-RL shifts the human burden from algorithm to task design.
In this work we automate the process of task design, devising a meta-learning algorithm
that does not require manual design of meta-training tasks.
We propose a family of unsupervised meta-RL algorithms
based on the insight that task proposals based on mutual information can be used to train optimal meta learners. 
Experimentally, our unsupervised meta-RL algorithm, which does not require manual task design, substantially improves on learning from scratch, and is competitive with supervised meta-RL approaches on benchmark~tasks.
\end{abstract}

\vspace{-2.5em}
\section{Introduction}
\label{sec:introduction}


Reusing past experience for faster learning of new tasks is a key challenge for machine learning. Meta-learning methods achieve this by using past experience to explicitly optimize for rapid adaptation~\citep{mishra2017simple, snell2017prototypical, schmidhuber1987evolutionary,finn2017model, gupta2018meta, wang2016learning, al2017continuous}. In the context of reinforcement learning (RL), meta-reinforcement learning (meta-RL) algorithms can learn to solve new RL tasks more quickly through experience on past tasks~\citep{duan2016rl, gupta2018meta, finn2017model}. Typical meta-RL algorithms assume the ability to sample from a pre-specified task distribution, and these algorithms learn to solve new tasks \emph{drawn from this distribution} very quickly. However, specifying a task distribution is tedious and requires a significant amount of supervision~\citep{oneshotimitation, duan2016rl} that may be difficult to provide for large, real-world problem settings. 
The performance of meta-learning algorithms critically depends on the meta-training task distribution, and meta-learning algorithms generalize best to new tasks which are drawn from the same distribution as the meta-training tasks~\citep{finn2017meta}.
In effect, meta-RL offloads the design burden from algorithm design to task design. While meta-RL acquires representations for fast adaptation to the specified task distribution, specifying this task distribution is often tedious and challenging. 
Can we automate the process of task design, thereby doing away with human supervision entirely?

In this paper, we take a step towards \emph{unsupervised} meta-RL: meta-learning from a task distribution that is acquired automatically, rather than requiring manual design of the meta-training tasks.
While unsupervised meta-RL does not make any assumptions about the reward functions on which it will be evaluated at test time, it does assume that the environment dynamics remain the same. This allows an unsupervised meta-RL agent to utilize environment interactions to meta-train a model that is optimized to be effective for learning from previously unseen reward functions in that environment at meta-test time.
Our method can also be thought of as automatically acquiring an \emph{environment-specific learning procedure} for deep neural network policies, somewhat related to data-driven initialization procedures explored in supervised learning~\citep{krahenbuhl2015data, hsu2018unsupervised}.

The primary contribution of our work is a framework for unsupervised meta-RL. We describe a family of unsupervised meta-RL algorithms and provide analysis to show that unsupervised meta-RL methods based on mutual information can be optimal, in a minimax sense.
Our experiments shows that, for a variety of robotic control tasks, unsupervised meta-RL can effectively acquire RL procedures. These procedures not only learn faster than standard RL approaches that learn from scratch, but also outperform prior methods that do pure exploration and then fine-tuning at test time. Our results even approach the performance of an oracle method that relies on hand-designed task distributions.

\vspace{-0.7em}
\section{Related Work}
\label{sec:related-work}
\vspace{-0.3em}


Our work lies at the intersection of meta-RL, goal generation, and unsupervised exploration.
Meta-learning algorithms use data from multiple tasks to learn how to learn, acquiring rapid adaptation procedures from experience~\citep{schmidhuber1987evolutionary,naik,thrun,bengiobengio1,hochreiter,mann,learntolearnbygdbygd,hugo,finn2017model,metanetworks}.
These approaches have been extended into the setting of RL~\citep{duan2016rl,wang2016learning,finn2017model,sung2017learning,gupta2018meta, gmps, houthooft2018evolved,bradlymetaexplore,rakelly2019efficient, nagabandi2018learning}.
In practice, the performance of meta-learning algorithms depends on the user-specified meta-training task distribution. 
We aim to lift this limitation and provide a general recipe for avoiding manual task engineering for meta-RL.
A handful of prior meta-learning methods have used self-proposed task distributions for learning supervised learning procedures~\citep{hsu2018unsupervised, antoniou2019assume, lin2019learning, ji2019unsupervised}. In contrast, our work deals with the RL setting, where the environment dynamics provides a rich inductive bias that our meta-learner can exploit.
In the RL setting, task distributions can be obtained in a variety of ways, including adversarial goal generation~\citep{sukhbaatar2017intrinsic,held2017automatic}, information-theoretic methods~\citep{gregor2016variational,eysenbach2018diversity, co2018self, achiam2018variational}.
The most similar work is~\citet{jabri2019unsupervised}, which also considers the unsupervised application of meta-learning to RL tasks. We build upon this work by proving that an optimal meta-learner can be acquired using mutual information-based task proposal.

Exploration methods that seek out novel states are also closely related to goal generation methods~\citep{deepakcuriosity, schmidhubercuriosity, countbased, bootstrapDQN, stadie2015incentivizing},
but do not by themselves aim to generate new tasks or learn to adapt more quickly to new tasks, only to achieve wide coverage of the state space. 
Model-based RL methods~\citep{deisenroth2011pilco,chua2018deep, srinivas2018universal, nagabandi2018neural,finn2017deep, atkeson1997comparison} use unsupervised experience to learn a dynamics model but do not learn how to efficiently use this model to explore to solve new tasks.

\begin{figure}[!t]
    \centering
    \includegraphics[width=\linewidth]{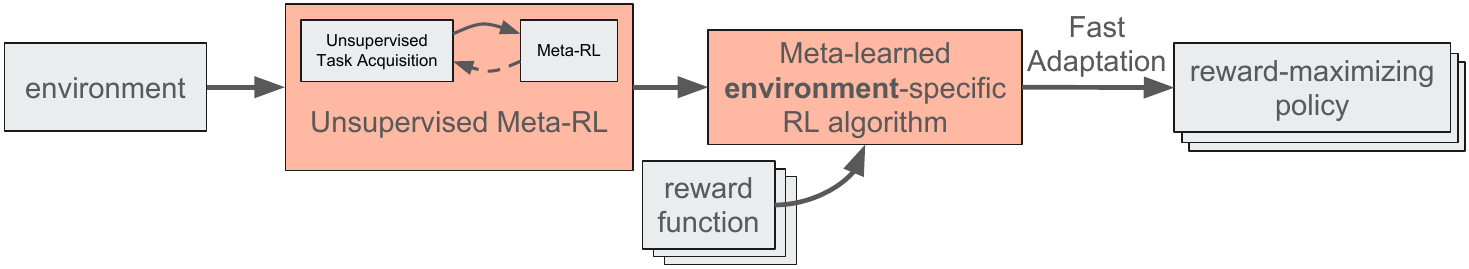}
    \vspace{-2.0em}
    \caption{\textbf{Unsupervised meta-reinforcement learning}: Given an environment, unsupervised meta-RL produces an environment-specific learning algorithm that quickly acquire new policies that maximize any task reward function.
    }
    \vspace{-1.5em}
    \label{fig:mud-rl}
\end{figure}

Goal-conditioned RL~\citep{schaul2015universal, andrychowicz2017hindsight,pong2018temporal} is also related to our work, and our analysis will study this special case first before generalizing to the general case of arbitrary tasks. As we discuss in Section~\ref{sec:theory_general}, goal-reaching itself is not enough, as goal-reaching agents are not optimized to efficiently explore to determine which goal they should reach, relying instead on a hand-specified goal parameterization that doesn't allow these algorithms to work with arbitrary reward functions.

\vspace{-0.7em}
\section{Unsupervised Meta-RL}
\vspace{-0.3em}

We consider the problem of \emph{learning} a reinforcement learning algorithm that can quickly solve new tasks in a given environment. This meta-RL process could, for example, tune the hyperparameters of another RL algorithm, or could replace the RL update rule itself with a learned update rule.
Unlike prior work, we aim to do so without depending on any human supervision or information about the tasks that will be provided for meta-testing.
A task reward is provided at meta-test time, and the learned RL procedure should adapt to this task reward as quickly as possible.
We assume that all test-time tasks have the same dynamics, and differ only in their reward functions.
Our algorithm will therefore need to utilize unsupervised environment interaction to learn an RL algorithm.
\emph{In effect, the dynamics themselves will be the supervision for our learning algorithm.}

We formalize the meta-training setting as a controlled Markov process (CMP) -- a Markov decision process without a reward function, $C = (S, A, P, \gamma, \rho)$, with state space $S$, action space $A$, transition dynamics $P$, discount factor $\gamma$ and initial state distribution $\rho$. The CMP, along with a reward function $r$, produces a Markov decision processes $M = (S, A, P, \gamma, \rho, r)$.
We define a learning algorithm $f: \gD \rightarrow \pi$ as a function that takes as input a dataset of experience from the MDP, $\gD = \{(s_i, a_i, r_i, s_i')\} \sim M$, and outputs a policy $\pi(a \mid s)$.
Evaluation of the learning procedure $f$ is carried out over a handful of episodes. In episode $i$, the learning procedure $f$ observes all previous data $\{\tau_1, \cdots, \tau_{i-1}\}$ and outputs a policy to be used in iteration $i$. We evaluate the learning procedure $f$ by summing its cumulative reward across iterations:
\vspace{-0.7em}
\begin{equation*}
    \vspace{-1.2em}
    R(f, r_z) = \sum_i \E_{\substack{\pi = f(\{\tau_1, \cdots, \tau_{i-1}\}) \\ \tau \sim \pi}} \left[\sum_t r_z(s_t, a_t) \right]
\end{equation*}
Our aim is to take this CMP and produce an environment-specific learning algorithm $f$ that can quickly learn an optimal policy $\pi_r^*(a \mid s)$ for \emph{any} reward function $r$. We refer to this problem as \emph{unsupervised meta-RL}, and illustrate the problem setting in  Fig.~\ref{fig:mud-rl}.

We now sketch a recipe for unsupervised meta-RL, analyze when this recipe is optimal, and then instantiate a practical approximation to this theoretically-motivated approach by building upon known meta-learning algorithms and unsupervised exploration methods.

\subsection{A General Recipe}
\label{sec:recipe}

To construct an unsupervised meta-RL algorithm, we leverage the insight that, to acquire a fast learning algorithm without task supervision, we can simply leverage standard meta-learning techniques, but with unsupervised task proposal mechanisms. Our unsupervised meta-RL framework therefore consists of a task proposal mechanism and a meta-learning method. For reasons that will become more apparent later, we will define the task distribution as a mapping from a latent variable $z \sim p(z)$ to a reward function $r_z(s,a) : \mathcal{S} \times \mathcal{A} \rightarrow \mathbb{R}$\footnote{In most cases $p(z)$ is chosen to be a uniform categorical so it is not challenging to specify}. That is, for each value of the random variable $z$, we have a different reward function $r_z(s,a)$. Under this formulation, learning a task distribution amounts to optimizing a parametric form for the reward function $r_z(s,a)$ that maps each $z \sim p(z)$ to a different reward function. The choice of this parametric form represents an important design decision for an unsupervised meta-learning method, and the resulting set of tasks is often referred to as a task or goal proposal procedure. In the following section, we will discuss a theoretical framework that allows us to make this choice in the following section so as to minimize worst case regret of the subsequently meta-learned learning algorithm $f$.

The second component is the meta-learning
algorithm, which takes the family of reward functions induced by $p(z)$ and $r_z(s,a)$, along with the associated CMP, and meta-learns an RL algorithm $f$ that can quickly adapt to any task from the task distribution defined by $p(z)$ and $r_z(s,a)$ in the given CMP. The meta-learned algorithm $f$ can then learn new tasks quickly at meta-test time, when a user-specified reward function is actually provided. Fig.~\ref{fig:mud-rl} summarizes this generic design for an unsupervised meta-RL algorithm.

The ``no free lunch theorem''~\citep{wolpert1995no, Whitley05complexitytheory} might lead us to expect that a truly generic approach to proposing a task distribution would not yield a learning procedure $f$ that is effective on any real tasks. However, \emph{the assumption that the dynamics remain the same across tasks affords us an inductive bias with which we pay for our lunch.}
In the following sections, we will discuss how to formulate acquiring the optimal unsupervised learning procedure, which minimizes regret on new meta-test tasks in the absence of any prior knowledge.
Since our analysis will focus on a restricted class of learning procedures, our results are lower bounds for the performance of general learning procedures.
We first define an optimal meta-learner and then show how we can train one without requiring task distributions to be hand-specified. 


\subsection{Optimal Meta-Learners}
\label{sec:optimalmetalearner}

We begin our analysis by considering the optimal learning procedure when the task distribution is known. For a task distribution $p(r_z)$, the optimal learning procedure $f^*$ is given by
\vspace{-1em}
\begin{equation*}
\vspace{-1em}
    f^* \triangleq \argmax_f \E_{p(r_z)} \left[R(f, r_z) \right].
\end{equation*}
Other learning procedures $f$ may achieve lower reward, and we define the regret incurred by using a suboptimal learning procedure as the difference in expected reward, compared with the optimal learning procedure:
\begin{equation*}
    \textsc{Regret}(f, p(r_z)) \triangleq \E_{p(r_z)} \left[R(f^*, r_z) \right] - \E_{p(r_z)} \left[R(f, r_z) \right].
\end{equation*}
Minimizing this regret is equivalent to maximizing the expected reward objective used by most meta-RL methods~\citep{finn2017model,duan2016rl}.
Note that different task distributions $p(r_z)$ will have different optimal learning procedures $f^*$.
For example, the optimal behavior for manipulation tasks involves moving a robot's arms, while the optimal behavior for locomotion tasks involves moving a robot's legs.
Therefore, $f^*$ depends on $p(r_z)$.
We next define the notion of an optimal \emph{unsupervised} meta-learner, which does not require prior knowledge of $p(r_z)$.

In unsupervised meta-reinforcement learning, the reward distribution $p(r_z)$ is unknown. In this setting, we evaluate a learning procedure $f$ based on its regret against the worst-case task distribution for CMP $C$:
\begin{equation}
    \textsc{Regret}_{\textsc{WC}}(f, C) = \max_{p(r_z)} \textsc{Regret}(f, p(r_z)). \label{eq:minimax}
\end{equation}
For a CMP $C$, we define the optimal unsupervised learning procedure as follows:
\begin{definition}
The optimal unsupervised learning procedure $f_C^*$ for a CMP $C$ is defined as
\vspace{-0.5em}
\begin{equation*}
\vspace{-1em}
    f_C^* \triangleq \argmin_f \textsc{Regret}_{\textsc{WC}}(f, C).
\end{equation*}
\end{definition}
Note the optimal unsupervised learning procedure may be different for different CMPs. We can also define the optimal unsupervised \emph{meta-learning} algorithm $\gF*$, which takes as input a CMP $C$ and returns the optimal unsupervised learning procedure $f_C^*$ for that CMP:
\begin{definition}
The optimal unsupervised meta-learner $\gF^*(C) = f_C^*$ is a function that takes as input a CMP $C$ and outputs the corresponding optimal unsupervised learning procedure $f_C^*$:
\vspace{-0.5em}
\begin{equation*}
\vspace{-1em}
    \gF^* \triangleq \argmin_{\gF} \textsc{Regret}_{\textsc{WC}}(\gF(C), C)
\end{equation*}
\end{definition}
Note that the optimal unsupervised meta-learner $\gF^*$ is universal -- it does not depend on any particular task distribution, or any particular CMP.
The next sections discuss how to find the minimax learning procedure, which minimizes the worst-case regret (Eq.~\ref{eq:minimax}).

\subsection{Special Case: Goal-Reaching Tasks}
\label{sec:goalreaching}

We start by deriving an optimal unsupervised meta-learner for the special case where all tasks are assumed to be goal state reaching tasks, and then generalize this approach to solve arbitrary tasks in Section~\ref{sec:theory_general}.  We restrict our analysis to CMPs with deterministic dynamics, and consider episodes with finite horizon $T$ and a discount factor of $\gamma = 1$. Each tasks corresponds to reaching a goal states $s_g$ at the last time step in the episode, so the reward function is
\vspace{-0.5em}
\begin{equation*}
\vspace{-0.5em}
    r_g(s_t) \triangleq \mathbbm{1}(t = T) \cdot \mathbbm{1}(s_t = g).
\end{equation*}
We first derive the optimal learning procedure for the case where $p(s_g)$ is known, and then derive the optimal procedure for the case where $p(s_g)$ is unknown.

\subsubsection{The Optimal Learning Procedure for Known~$p(s_g)$} 
In the case of goal reaching tasks, the optimal fast learning procedure $f$ searches through potential goal states until it finds the goal and then navigates to that goal state in all subsequent episodes.
Define $f_\pi$ as the learning procedure that uses policy $\pi$ to explore until the goal is found, and then always returns to the goal state.
We will restrict our attention to the set of learning procedures $\gF_\pi \triangleq \{f_\pi\}$ constructed in this fashion, so our theoretical results will be lower bound on the performance of arbitrary learning procedures.
The learning procedure $f_\pi$
incurs one unit of regret for each step before it has found the goal, and zero regret afterwards. The expected cumulative regret is therefore the expectation of the hitting time.
To compute the expected hitting time, we define $\rho_\pi^T(s)$ as the probability that policy $\pi$ visits state $s$ at time step $t = T$. If $s_g$ is the true goal, then the event that the policy $\pi$ reaches $s_g$ at the final step of an episode is a Bernoulli random variable with parameter $p = \rho_\pi^T(s_g)$. Thus, the expected hitting time of this goal state is
\vspace{-0.5em}
\begin{equation*}
\vspace{-0.5em}
    \textsc{HittingTime}_\pi(s_g) = \frac{1}{\rho_\pi^T(s_g)}.
\end{equation*}
The regret of the learning procedure $f_\pi$ is 
\vspace{-0.7em}
\begin{multline}
\vspace{-0.5em}
    \textsc{Regret}(f_\pi, p(r_g)) = \int \textsc{HittingTime}_\pi(s_g) p(s_g) d s_g \\ = \int \frac{p(s_g)}{\rho_\pi^T(s_g)} d s_g. \label{eq:goal-reaching-regret}
\end{multline}
To now compute the \emph{optimal} learning procedure $f_\pi$,
we can minimize the regret in Equation~\ref{eq:goal-reaching-regret} w.r.t.\ the marginal distribution $\rho_\pi^T$. Using the calculus of variations (for more details refer to Appendix C in~\citet{lee2019smm}), the exploration policy for the optimal meta-learner, $\pi^*$, satisfies:
\begin{equation}
    \rho_{\pi^*}^T(s_g) = \frac{\sqrt{p(s_g)}}{\bigintsss \sqrt{p(s_g')} d s_g'}. \label{eq:opt-policy}
\end{equation}
Thus, when the goal sampling distribution $p(s_g)$ is known, the optimal learning procedure is obtained by finding $\pi^*$ satisfying Eq.~\ref{eq:opt-policy} and then using $f_{\pi^*}$ as the learning procedure. The next section considers the case where $p(s_g)$ is not known.

\subsubsection{The Optimal Unsupervised Learning Procedure for Goal Reaching Tasks}

In the case of goal-reaching tasks where the goal distribution $p(s_g)$ is not known, the optimal unsupervised learning procedure can be constructed from a policy with a uniform marginal state distribution (proof in Appendix~\ref{sec:proofs}):
\begin{lemma} \label{lemma:uniform}
Let $\pi$ be a policy for which $\rho_\pi^T(s)$ is uniform. Then $f_\pi$ is has lowest worst-case regret among learning procedures in $\gF_\pi$.
\end{lemma}

One route for constructing this optimal unsupervised learning procedure is to first acquire a policy $\pi$ for which $\rho_\pi^T(s)$ is uniform and then return $f_\pi$. However, finding such a policy $\pi$ is challenging, especially in high-dimensional state spaces and in the absense of resets. Instead, we will take an alternate route, acquiring $f_\pi$ directly without every computing $\pi$. In addition to sidestepping the requirement of computing $\pi$, this approach will also have the benefit of generalizing beyond goal-reaching tasks to arbitrary task distributions.

Our approach for directly computing the optimal unsupervised learning procedure hinges on the observation that the optimal unsupervised learning procedure is the optimal (supervised) learning procedure for goals proposed from a uniform distribution. Thus, the optimal unsupervised learning procedure will come not as a result of a careful construction, but rather as the output of the an optimization procedure (i.e., meta-learning).
Thus, we can obtain the optimal unsupervised learning procedure by applying a meta-learning algorithm to a task distribution that samples goals uniformly. To ensure that the resulting learning procedure $f$ lies within the set $\gF_\pi$, we will only consider ``memoryless'' meta-learning algorithms that maintain no internal state before the true goal is found.\footnote{MAML satisfies this requirement, as the internal parameters are updated by policy gradient, which is zero because the reward is zero before the true goal is found.}
While sampling goals uniform is itself a challenging problem, we can use the same trick as before: instead of constructing this uniform goal distribution directly, we instead find an optimization problem for which the solution is to sample goals uniformly.

The optimization problem that we use will involve two latent variables, the final state $s_T$ and an auxiliary latent variable $z$ sampled from a prior $\mu(z)$. The optimization problem will be to find a conditional distribution $\mu(s_T \mid z)$ such that the mutual information between $z$ and $s_T$ is optimized:
\begin{equation}
    \max_{\mu(s_T \mid z)} I_\mu(s_T; z) \label{eq:diayn}
\end{equation}
The conditional distribution $\mu(s_T \mid z)$ that optimizes Equation~\ref{eq:diayn} is one with a uniform marginal distribution over terminal states (proof in Appendix~\ref{sec:proofs}):
\begin{lemma} \label{lemma:MImax}
Assume there exists a conditional distribution $\mu(s_T \mid z)$ satisfying the following two properties:
\begin{enumerate}[topsep=1pt,itemsep=1pt]
    \item The marginal distribution over terminal states is uniform: $\mu(s_T) = \int \mu(s_T \mid z) \mu(z) dz = \textsc{Unif}(\gS)$; and
    \item The conditional distribution $\mu(s_T \mid z)$ is a Dirac: $\forall z, s_T \; \exists s_z \text{ s.t. } \mu(s_T \mid z) = \mathbbm{1}(s_T = s_z)$.
\end{enumerate}
Then any solution $\mu(s_T \mid z)$ to the mutual information objective (Eq.~\ref{eq:diayn}) satisfies the following:
\begin{equation*}
\mu(s_T) = \textsc{Unif}(\gS) \quad \text{and} \quad \mu(s_T \mid z) = \mathbbm{1}(s_T = s_z).
\end{equation*}
\end{lemma}

\vspace{-0.5em}
\subsubsection{Optimizing Mutual Information} 
To optimize the above mutual information objective, we note that a conditional distribution $\mu(s_T \mid z)$ can be defined implicitly via a latent-conditioned policy $\mu(a \mid s, z)$. This policy is \emph{not} a meta-learned model, but rather will become part of the task proposal mechanism. For a given prior $\mu(z)$ and latent-conditioned policy $\mu(a \mid s, z)$, the joint likelihood is 
\vspace{-0.5em}
\begin{equation*}
\vspace{-0.5em}
\mu(\tau, z) = \mu(z) p(s_1)\prod_t p(s_{t+1} \mid s_t, a_t) \mu(a_t \mid s_t, z),
\end{equation*}
and the marginal likelihood is simply given by
\vspace{-0.5em}
\begin{equation*}
\vspace{-0.5em}
\mu(s_T, z) = \int \mu(\tau, z) ds_1a_1 \cdots a_{T-1}.
\end{equation*}
The purpose of our repeated indirection now becomes clear: prior work~\citep{eysenbach2018diversity,achiam2018variational} has proposed efficient algorithms for maximizing the mutual information objective (Eq.~\ref{eq:diayn}) when the conditional distribution $\mu(s_T \mid z)$ is defined implicitly in terms of a latent-conditioned policy.
At this point, we finally can sample goals uniformly, by sampling $z \sim \mu(z)$ followed by $s_T \sim \mu(s_T \mid z)$. 

Recall that we wanted to obtain a uniform goal distribution so that we could apply meta-learning to obtain the optimal learning procedure. However, the input to meta-learning procedures is not a distribution over goals but a distribution over reward functions.
We then define our task proposal distribution $p(r_z)$ by sampling $z \sim p(z)$ and using the corresponding reward function $r_z(s_T, a_T) \triangleq \log p(s_T \mid z)$, resulting in a uniform distribution as described in Lemma 2. 

\subsection{General Case: Trajectory-Matching Tasks}
\label{sec:theory_general}

To extend the analysis in the previous section to the general case, and thereby derive a framework for optimal unsupervised meta-learning, we will consider ``trajectory-matching'' tasks. These tasks are a trajectory-based generalization of goal reaching: while goal reaching tasks only provide a positive reward when the policy reaches the goal state, trajectory-matching tasks only provide a positive reward when the policy executes the optimal trajectory. The trajectory matching case is more general
because, while trajectory matching can represent different goal-reaching tasks, it can also represent tasks that are not simply goal reaching, such as reaching a goal while avoiding a dangerous region or reaching a goal in a particular way. Moreover, the trajectory matching case is actually also a generalization of the typical reinforcement learning case with Markovian rewards, because any such task can be represented by a trajectory reaching objective as well. Please refer to Section~\ref{sec:generalreward} for a more complete discussion of the same. 

As before, we will restrict our attention to CMPs with deterministic dynamics. These non-Markovian tasks essentially amount to a problem where an RL algorithm must ``guess'' the optimal policy, and only receives a reward if its behavior is perfectly consistent with that optimal policy.

We will show that optimizing the mutual information between $z$ and \emph{trajectories} to obtain a task proposal distribution, and subsequently optimizing a meta-learner for this distribution will give us the optimal unsupervised meta-learner for this class of reward functions. We subsequently show that unsupervised meta-learning for the trajectory-matching task is at least as hard as unsupervised meta-learning for general tasks. As before, let us begin within an analysis of optimal meta-learners in the case where the distribution over trajectory matching tasks $p(\tau^*)$ is known, and subsequently direct our attention to formulating an optimal unsupervised meta-learner. 


\subsubsection{Optimal meta-learner for known $p(\tau^*)$}
Formally, we define a distribution of trajectory-matching tasks by a distribution over desired trajectories, $p(\tau^*)$. For each goal trajectory $\tau^*$, the corresponding trajectory-level reward function is
\vspace{-1em}
\begin{equation*}
\vspace{-0.5em}
    r_\tau^*(\tau) \triangleq \mathbbm{1}(\tau = \tau^*)
\end{equation*}
Analysis from Section~\ref{sec:goalreaching} can be repurposed here. As before, restrict our attention to learning procedures $f_\pi \in \gF_\pi$. After running the exploration policy to discover trajectories that obtain reward, the policy will deterministically keep executing the desired trajectory.  We can define the hitting time as the expected number of episodes to match the target trajectory:
\vspace{-0.5em}
\begin{equation*}
\vspace{-0.5em}
    \textsc{HittingTime}_\pi(\tau^*) = \frac{1}{\pi(\tau^*)}
\end{equation*}
We then define regret as the expected hitting time:
\begin{multline}
    \textsc{Regret}(f_\pi, p(r_\tau)) = \int \textsc{HittingTime}_\pi(\tau) p(\tau) d \tau) \\ = \int \frac{p(\tau)}{\pi(\tau)} d \tau.
\end{multline}
This definition of regret allows us to optimize for an optimal learning procedure, and we obtain an exploration policy for the optimal learning procedure satisfying the requirement
\vspace{-0.7em}
\begin{equation*}
\vspace{-0.5em}
    \pi^*(\tau) = \frac{\sqrt{p(\tau)}}{\int \sqrt{p(\tau')} d\tau'}.
\end{equation*}

\subsubsection{Optimal unsupervised learning procedure for trajectory-matching tasks}
As described in Section~\ref{sec:optimalmetalearner}, obtaining such a policy requires knowing the trajectory distribution $p(\tau)$, and we must resort to optimizing the worst-case regret. As argued in Lemma 1, the solution to this min-max optimization is a learning procedure which has an exploration policy that is uniform distribution over trajectories. 
\begin{lemma}
Let $\pi$ be a policy for which $\pi(\tau)$ is uniform. Then $f_\pi$ has lowest worst-case regret among learning procedures in $\gF_\pi$.
\end{lemma}
We can acquire an unsupervised meta-learner of this form by proposing and meta-learning on a task distribution that is uniform over trajectories. How might we actually propose a task distribution that is uniform over trajectories? As argued for the goal reaching case, we can do so by optimizing a \emph{trajectory-level} mutual information objective:
\vspace{-1em}
\begin{equation*}
\vspace{-0.5em}
    I(\tau; z) = \gH[\tau] - \gH[\tau \mid z]
\end{equation*}
The optimal policy for this objective has a uniform distribution over trajectories that, conditioned on a particular latent $z$, deterministically produces a single trajectory in a deterministic CMP. The analysis for the case of stochastic dynamics is more involved and is left to future work. By optimizing a task proposal distribution that maximizes trajectory-level mutual information, and subsequently performing meta-learning on the proposed tasks, we can acquire the optimal unsupervised meta-learner for trajectory matching tasks, under the definition in Section~\ref{sec:optimalmetalearner}. 

\subsubsection{Relationship to General Reward Maximizing Tasks}
\label{sec:generalreward}
Now that we have derived the optimal meta-learner for trajectory-matching tasks, we observe that trajectory-matching is a super-set of the problem of optimizing any possible Markovian reward function at test-time. For a given initial state distribution, each reward function is optimized by a particular trajectory. However, trajectories produced by a non-Markovian policy (i.e., a policy with memory) are not necessarily the unique optimum for any Markovian reward function. Let $R_\tau$ denote the set of trajectory-level reward functions, and $R_{s,a}$ denote the set of all state-action level reward functions. Bounding the worst-case regret on $R_\tau$ minimizes an upper bound on the worst-case regret on $R_{s, a}$:
\begin{equation*}
    \min_{r_\tau \in R_\tau} \E_\pi \left[r_\tau(\tau) \right] \le \min_{r \in R_{s, a}} \E_\pi \left[\sum_t r(s_t, a_t) \right] \quad \forall \pi. 
\end{equation*}
This inequality holds for all policies $\pi$, including the policy that maximizes the LHS.
While we aim to maximize the RHS, we only know how to maximize the LHS, which gives us a lower bound on the RHS.
This inequality holds for all policies $\pi$, so it also holds for the policy that maximizes the LHS.
In general, this bound is loose, because the set of all Markovian reward functions is smaller than the set of all trajectory-level reward functions (i.e., trajectory-matching tasks). However, this bound becomes tight when considering meta-learning on the set of all possible (non-Markovian) reward functions.

In the discussion of meta-learning thus far, we have restricted our attention to tasks where the reward is provided at the last time step $T$ of each episode and to the set of learning procedures $\gF_\pi$ that maintain no internal state before the true goal or trajectory is found. In this restricted setting case, the best that an optimal meta-learner can do is go directly to a goal or execute a particular trajectory at every episode according to the optimal exploration policy as discussed previously, essentially performing a version of posterior sampling.
In the more general case with arbitrary reward functions and arbitrary learning procedures, intermediate rewards along a trajectory may be informative, and the optimal exploration strategy may be different from posterior sampling~\citep{promp, duan2016rl, wang2016learning}. 

Nonetheless, the analysis presented in this section provides us insight into the behavior of optimal meta-learning algorithms and allows us to understand the qualities desirable for unsupervised task proposals. The general proposed scheme for unsupervised meta-learning has a significant benefit over standard universal value function and goal reaching style algorithms: it can be applied to arbitrary reward functions going beyond simple goal reaching, and doesn't require the goal to be known in a parametric form beforehand. 

\vspace{-0.5em}
\subsection{Summary of Analysis}
Through our analysis, we introduced the notion of optimal meta-learners and analyze their exploration behavior and regret on a class of goal reaching problems. We showed that on these problems, when the test-time task distribution is unknown, the optimal meta-training task distribution for minimizing worst-case test-time regret is \emph{uniform} over the space of goals. We also showed that this optimal task distribution can be acquired by a simple mutual information maximization scheme. We subsequently extend the analysis to the more general case of matching arbitrary trajectories, as a proxy for the more general class of arbitrary reward functions. In the following section, we will discuss how we can derive a practical algorithm for unsupervised meta-learning from this analysis.

\subsection{A Practical Algorithm}

\setlength{\textfloatsep}{1pt}

Following the derivation in the previous section, we can instantiate a practical unsupervised meta-RL algorithm by constructing a task proposal mechanism based on a mutual information objective.
A variety of different mutual information objectives can be formulated, including mutual information between single states and $z$~\citep{eysenbach2018diversity}, pairs of start and end states and $z$~\citep{gregor2016variational}, and entire trajectories and $z$~\citep{achiam2018variational,sharma2019dynamics, warde2018unsupervised}. 
We will use DIAYN and leave a full examination of possible mutual information objectives for future work.
\begin{algorithm}[t]
   \caption{Unsupervised Meta-RL Pseudocode}
   \label{alg:unsup-ml}
\begin{algorithmic}
   \STATE {\bfseries Input:} $\mathcal{M} \setminus R$, an MDP without a reward function
   \STATE $D_{\phi} \leftarrow \text{DIAYN}()$ or $D_{\phi} \leftarrow random$
   \WHILE{not converged}
      \STATE Sample latent task variables $z \sim p(z)$
      \STATE Define task reward $r_z(s)$ using $D_{\phi}(z|s)$
      \STATE Update $f$ using MAML with reward $r_z(s)$
   \ENDWHILE
   \STATE {\bfseries Return:} a learning algorithm $f: D_{\phi} \rightarrow \pi$
\end{algorithmic}
\end{algorithm}

DIAYN optimizes mutual information by training a \emph{discriminator} network $D_\phi(z | \cdot)$ that predicts which $z$ was used to generate the states in a given rollout according to a \emph{latent-conditioned policy} $\pi(a|s,z)$.
Our task proposal distribution is thus defined by $r_z(s, a) = \log(D_{\phi}(z|s))$.
The complete unsupervised meta-learning algorithm is as follows: first, we acquire $r_z(s,a)$ by running DIAYN, which learns $D_{\phi}(z|s)$ and a latent-conditioned policy $\pi(a|s,z)$ (which is discarded). Then, we use $z \sim p(z)$ to propose tasks $r_z(s,a)$ to a standard meta-RL algorithm. This meta-RL algorithm uses the proposed tasks to learn how to learn, acquiring a fast learn algorithm $f$ which can then learn new tasks quickly. While, in principle, any meta-RL algorithm could be used, we use MAML~\citep{finn2017model} as our meta-learning algorithm. Note that the learning algorithm $f$ returned by MAML is defined simply as running gradient descent using the initial parameters found by MAML as initialization, as discussed in prior work ~\cite{chelseauniversality}. The method is summarized in Algorithm~\ref{alg:unsup-ml}.

In addition to mutual information maximizing task proposals, we will also consider random task proposals, where we also use a discriminator as the reward, according to $r(s,z) = \log D_{\phi_{\text{rand}}}(z|s)$, but where the parameters $\phi_{\text{rand}}$ are chosen randomly (i.e., a random weight initialization for a neural network). While such random reward functions are not optimal, we find that they can surprisingly be used to acquire useful task distributions for simple tasks, though they are not as effective as the tasks become more complicated.

\section{Experimental Evaluation}
\label{sec:experiments}

In our experiments, we aim to understand whether unsupervised meta-learning as described in Section~\ref{sec:recipe} can provide us with an accelerated RL procedure on new tasks. Whereas standard meta-learning requires a hand-specified task distribution at meta-training time, unsupervised meta-learning learns the task distribution through unsupervised interaction with the environment. A fair baseline that likewise uses requires \emph{no reward supervision} at training time, and only uses rewards at test time, is learning via RL from scratch without any meta-learning. As an upper bound, we include the \emph{unfair} comparison to a standard meta-learning approach, where the meta-training distribution is manually designed. This method has access to a hand-specified task distribution that is not available to our method. We evaluate two variants of our approach: (a) task acquisition based on DIAYN followed by meta-learning using MAML, and (b) task acquisition using a randomly initialized discriminator followed by meta-learning using MAML.

\begin{figure*}[!t]
    \centering
        \begin{subfigure}[b]{0.25\linewidth}
          \centering\includegraphics[width=\textwidth]{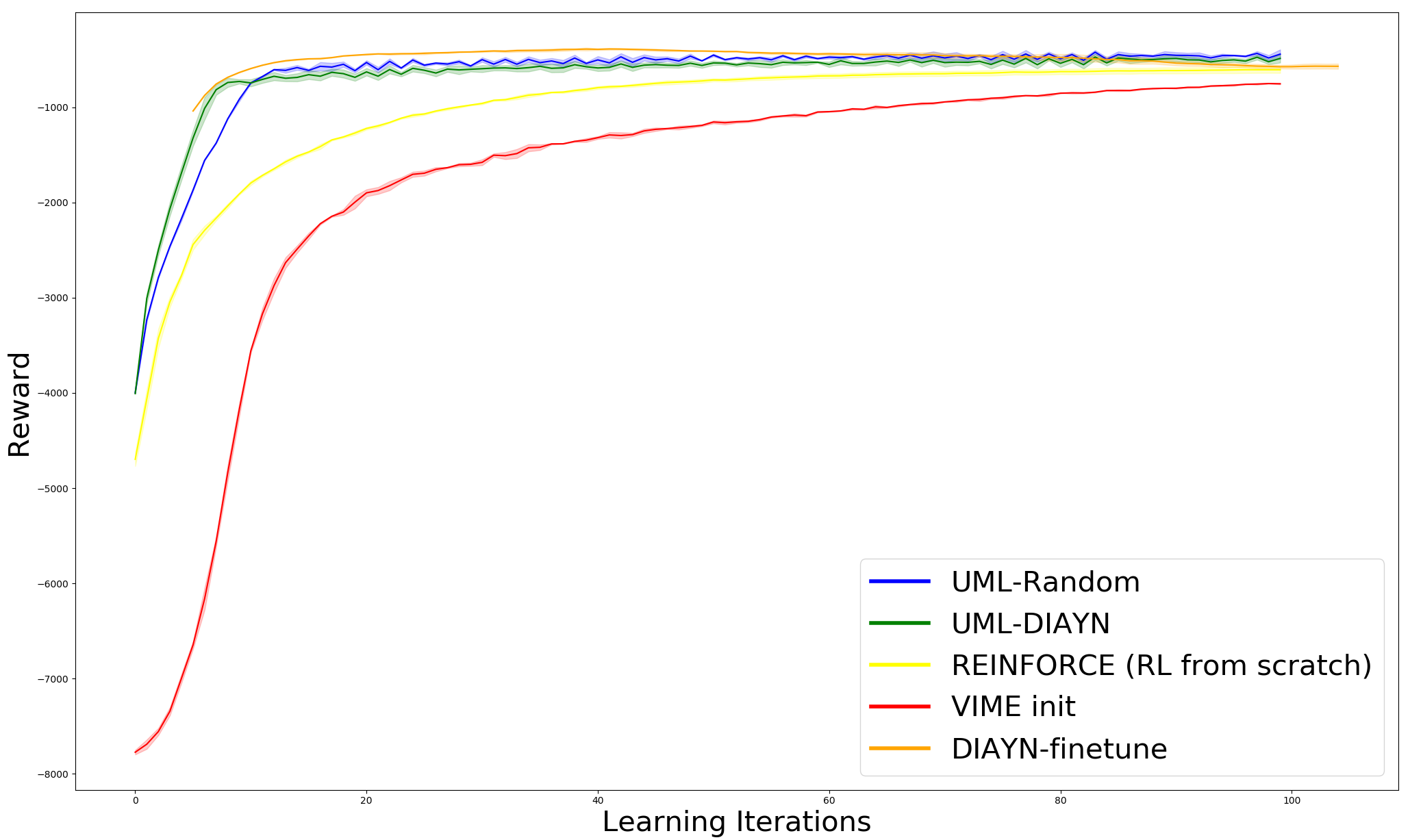}
          \caption*{2D navigation}
        \end{subfigure}%
        ~
        \begin{subfigure}[b]{0.25\linewidth}
          \centering\includegraphics[width=\textwidth]{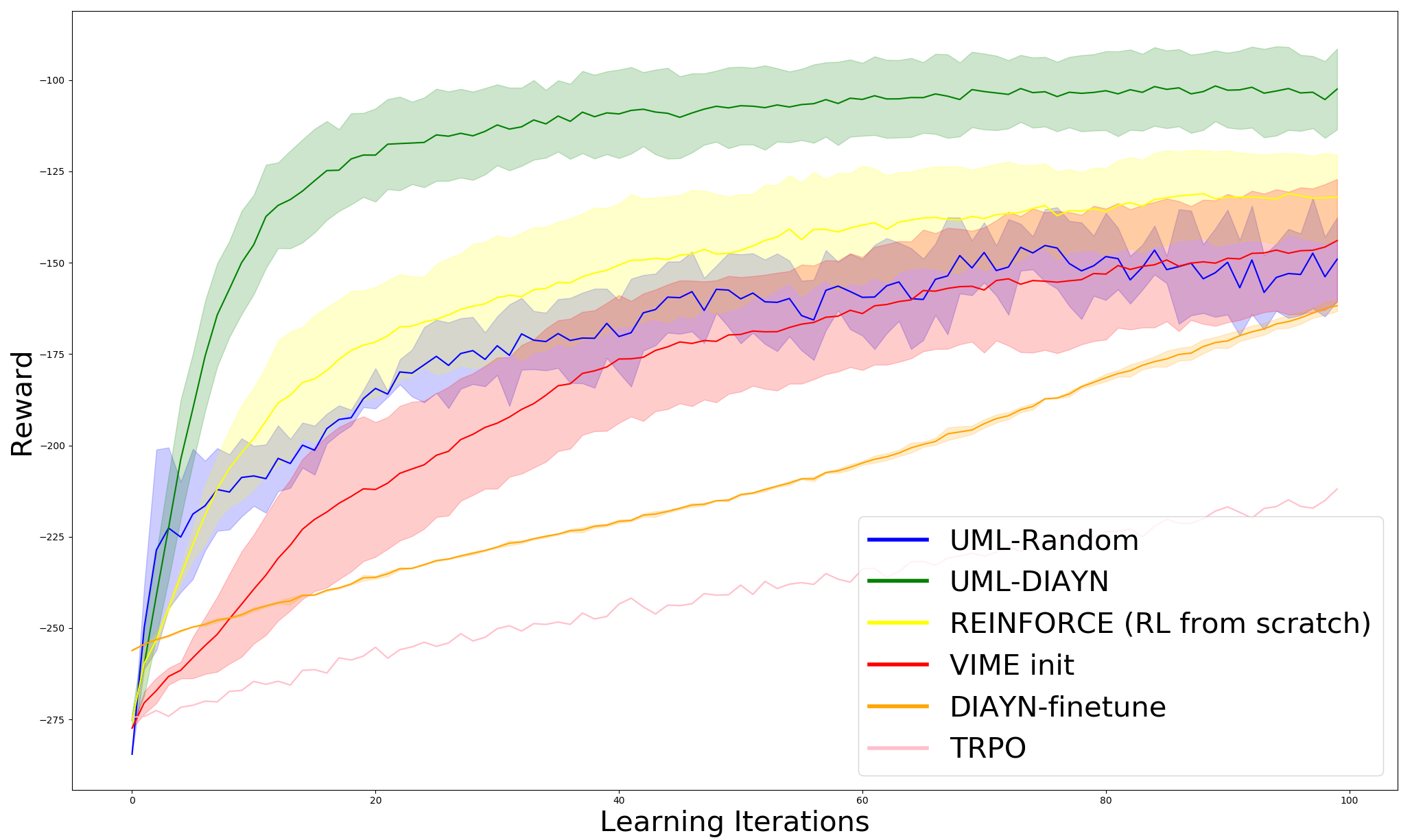}
          \caption*{Half-Cheetah}
       \end{subfigure}%
       ~
       \begin{subfigure}[b]{0.25\linewidth}
          \centering\includegraphics[width=\textwidth]{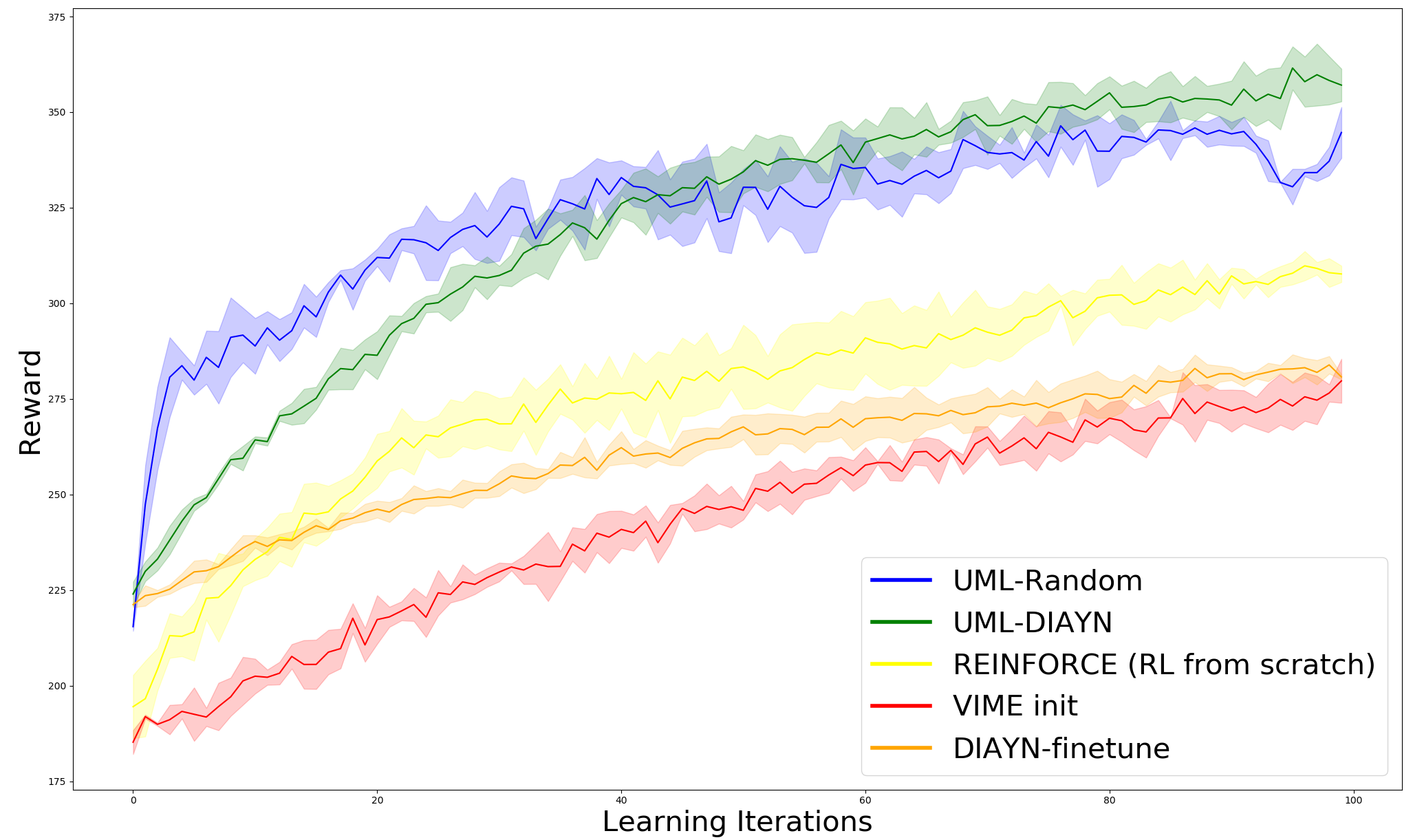}
          \caption*{Ant}
        \end{subfigure}
    \caption{\footnotesize
    \textbf{Unsupervised meta-learning accelerates learning}: After unsupervised meta-learning, our approach (UML-DIAYN and UML-RANDOM) quickly learns a new task significantly faster than learning from scratch, especially on complex tasks. Learning the task distribution with DIAYN helps more for complex tasks. Results are averaged across 20 evaluation tasks, and 3 random seeds for testing. UML-DIAYN and random also significantly outperform learning with DIAYN initialization or VIME.
    \label{fig:reward}}
    \vspace{-0.3em}
\end{figure*}

\begin{figure*}[!t]
    \centering
        \begin{subfigure}[b]{0.25\linewidth}
          \centering\includegraphics[width=\textwidth]{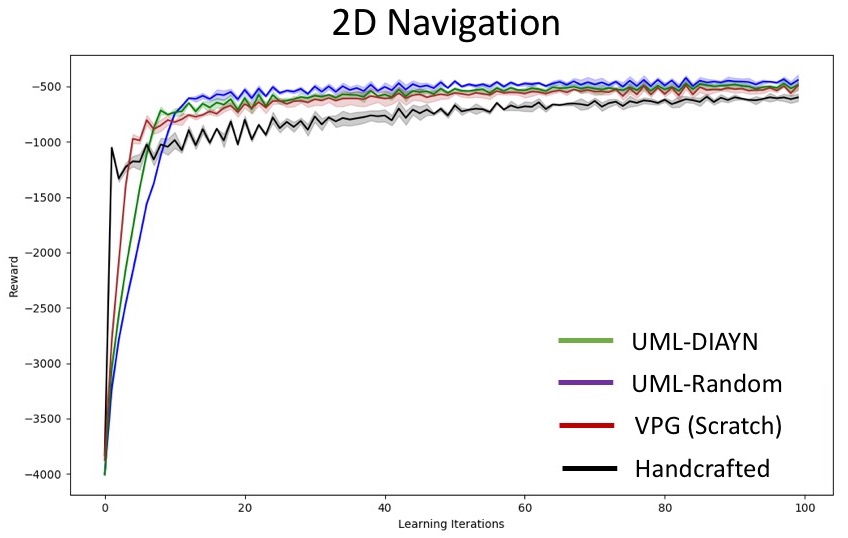}
          \caption*{2D Navigation}
        \end{subfigure}%
        ~
        \begin{subfigure}[b]{0.25\linewidth}
          \centering\includegraphics[width=\textwidth]{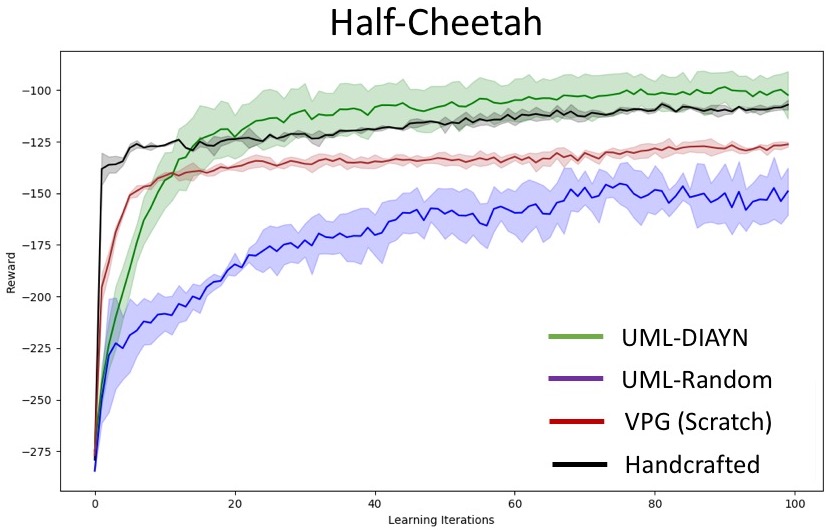}
          \caption*{Half-Cheetah}
       \end{subfigure}%
       ~
        \begin{subfigure}[b]{0.25\linewidth}
          \centering\includegraphics[width=\textwidth]{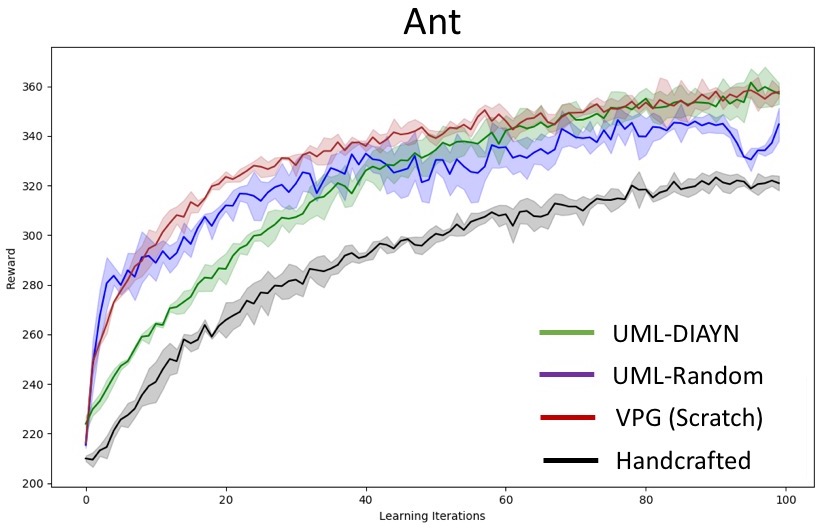}
          \caption*{Ant Navigation}
        \end{subfigure}
    \caption{ \textbf{Comparison with handcrafted tasks}: Unsupervised meta-learning (UML-DIAYN) is competitive with meta-training on handcrafted reward functions (i.e., an oracle). A misspecified, handcrafted meta-training task distribution often performs worse, illustrating the benefits of learning the task distribution.
    \label{fig:handcrafted}}
    \vspace{-0.5em}
\end{figure*}

\subsection{Tasks and Implementation Details}
Our experiments study three simulated environments of varying difficulty: 2D point navigation, 2D locomotion using the ``HalfCheetah,'' and 3D locomotion using the ``Ant,'' with the latter two environments are modifications of popular RL benchmarks~\citep{duan2016benchmarking}.
While the 2D navigation environment allows for direct control of position, HalfCheetah and Ant can only control their center of mass via feedback control with high dimensional actions (6D for HalfCheetah, 8D for Ant) and observations (17D for HalfCheetah, 111D for Ant).

The evaluation tasks, shown in Figure~\ref{fig:train-distribution}, are similar to prior work~\citep{finn2017model,pong2018temporal}: 2D navigation and ant require navigating to goal positions, while the half cheetah must run at different goal velocities. These tasks are not accessible to our algorithm during meta-training. Please refer to Appendix C for details about hyperparameters for both MAML and DIAYN. 

\subsection{Fast Adaptation after Unsupervised Meta RL}

The comparison between the two variants of unsupervised meta-learning and learning from scratch is shown in Figure~\ref{fig:reward}.
We also add a comparison to VIME~\citep{vime}, a standard novelty-based exploration method, where we pretrain a policy with the VIME reward and then finetune it on the meta-test tasks.
In all cases, the UML-DIAYN variant of unsupervised meta-learning produces an RL procedure that outperforms RL from scratch and VIME-init, suggesting that unsupervised interaction with the environment and meta-learning is effective in producing environment-specific but task-agnostic priors that accelerate learning on new, previously unseen tasks. The comparison with VIME shows that the speed of learning is not just about exploration but is indeed about fast adaptation. 
In our experiments thus far, UML-DIAYN always performs better than learning from scratch, although the benefit varies across tasks depending on the actual performance of DIAYN. We also perform significantly better than a baseline of simply initializing from a DIAYN trained contextual policy, and then finetuning the best skill with the actual task reward.

Interestingly, in many cases (in Figure~\ref{fig:handcrafted}) the performance of unsupervised meta-learning with DIAYN matches that of the hand-designed task distribution. We see that on the 2D navigation task, while handcrafted meta-learning is able to learn very quickly initially, it performs similarly after 100 steps. For the cheetah environment as well, handcrafted meta-learning is able to learn very quickly to start off, but is quickly matched by unsupervised meta-RL with DIAYN. On the ant task, we see that hand-crafted meta-learning does do better than UML-DIAYN, likely because the task distribution is challenging, and a better unsupervised task proposal algorithm would improve performance. 

The comparison between the two unsupervised meta-learning variants is also illuminating: while the DIAYN-based variant of our method generally achieves the best performance, even the random discriminator is often able to provide a sufficient diversity of tasks to produce meaningful acceleration over learning from scratch in the case of 2D navigation and ant. This result has two interesting implications. First, it suggests that unsupervised meta-learning is an effective tool for learning an environment prior. Although the performance of unsupervised meta-learning can be improved with better coverage using DIAYN (as seen in Figure~\ref{fig:reward}), even the random discriminator version provides competitive advantages over learning from scratch. Second, the comparison provides a clue for identifying the source of the structure learned through unsupervised meta-learning: though the particular task distribution has an effect on performance, simply interacting with the environment (without structured objectives, using a random discriminator) already allows meta-RL to learn effective adaptation strategies in a given environment.

\section{Discussion and Future Work}
\label{sec:limitations}

We presented an unsupervised approach to meta-RL, where meta-learning is used to acquire an efficient RL procedure without requiring hand-specified task distributions. This approach accelerates RL without relying on the manual supervision required for conventional meta-learning algorithms. We provide a theoretical derivation that argues that task proposals based on mutual information maximization can provide a minimum worst-case regret meta-learner, under certain assumptions. Our experiments indicate unsupervised meta-RL can accelerate learning on a range of tasks.


Our approach also opens a number of questions about unsupervised meta-learning algorithms. One limitation of our analysis is that it only considers deterministic dynamics, and only considers task distributions where posterior sampling is optimal. Extending our analysis to stochastic dynamics and more realistic task distributions may allow unsupervised meta-RL to acquire learning algorithms that can more effectively solve real-world tasks.



{\footnotesize
\bibliography{references}
\bibliographystyle{iclr2020_conference}
}

\clearpage
\appendix

\section{Proofs}
\label{sec:proofs}

\textbf{Lemma 1} Let $\pi$ be a policy for which $\rho_\pi^T(s)$ is uniform. Then $\pi$ has lowest worst-case regret.

\begin{proof}[Proof of Lemma~\ref{lemma:uniform}]
To begin, we note that all goal distributions $p(s_g)$ have equal regret for policies where $\rho_\pi^T(s) = 1 / |\gS|$ is uniform:
\begin{equation*}
    \textsc{Regret}_p(\pi) = \int \frac{p(s_g)}{\rho_\pi^T(s_g)} ds_g = \int \frac{p(s_g)}{1/|\gS|} ds_g = |\gS|
\end{equation*}
Now, consider a policy $\pi'$ for which $\rho_\pi^T(s)$ is not uniform. For simplicity, we will assume that the argmin is unique, though the proof holds for non-unique argmins as well. The worst-case goal distribution will choose the state $s^-$ where that the policy is least likely to visit:
\begin{equation*}
    p^-(s_g) \triangleq \mathbbm{1}(s_g = \argmin_s \rho_\pi^T(s))
\end{equation*}
Thus, the worst-case regret for policy $\pi'$ is strictly greater than the regret for a uniform $\pi$:
\begin{multline}
     \max_p \textsc{Regret}_p(\pi) = \textsc{Regret}_{p^-}(\pi) \\ =  \int \frac{\mathbbm{1}(s_g = \argmin_s \rho_\pi^T(s))}{\rho_\pi^T(s_g)} ds_g \\ = \frac{1}{\min_s \rho_{\pi'}^T(s)} > |\gS|
\end{multline}
Thus, a policy $\pi'$ for which $\rho_\pi^T$ is non-uniform cannot be minimax, so the optimal policy has a uniform marginal $\rho_\pi^T$.
\end{proof}

\textbf{Lemma 2}: Mutual information $I(s_T; z)$ is maximized by a task distribution $p(s_g)$ which is uniform over goal states.

\begin{proof}[Proof of Lemma~\ref{lemma:MImax}]
We define a latent variable model, where we sample a latent variable $z$ from a uniform prior $p(z)$ and sample goals from a conditional distribution $p(s_T \mid z)$. To begin, note that the mutual information can be written as a difference of entropies:
\begin{equation*}
    I_p(s_T; z) = \gH_p[s_T] - \gH_p[s_T \mid z]
\end{equation*}
The conditional entropy $\gH_p[s_T \mid z]$ attains the smallest possible value (zero) when each latent variable $z$ corresponds to exactly one final state, $s_z$. In contrast, the marginal entropy $\gH_p[s_T]$ attains the largest possible value ($\log |\gS|$) when the marginal distribution $p(s_T) = \int p(s_T \mid z) p(z) dz$ is uniform. Thus, a task uniform distribution $p(s_g)$ maximizes $I(s_T; z)$. Note that for any non-uniform task distribution $q(s_T)$, we have $\gH_q[s_T] < \gH_p[s_T]$. Since the conditional entropy $\gH_p[s_T \mid z]$ is zero, no distribution can achieve a smaller conditional entropy. This, for all non-uniform task distributions $q$, we have $I_q(s_T; z) < I_p(s_T; z)$. Thus, the optimal task distribution must be uniform.
\end{proof}

\section{Ablations}
\begin{figure}[!h]
  \centering
  \includegraphics[width=\columnwidth]{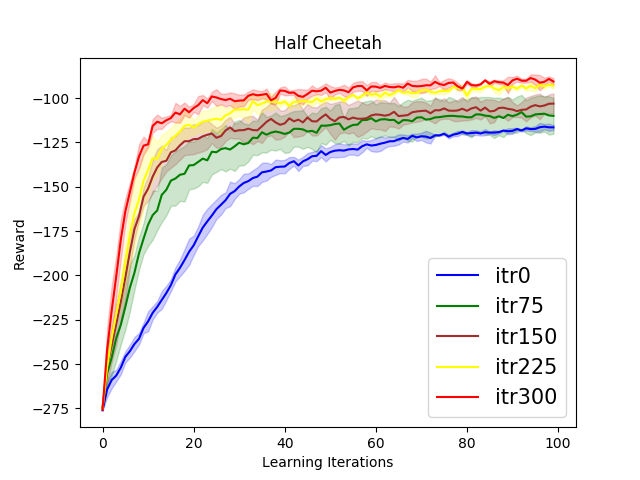}
  \vspace{-1em}
  \caption{\footnotesize Analysis of effect of additional meta-training on meta-test time learning of new tasks. For larger iterations of meta-trained policies, we have improved test time performance, showing that additional meta-training is beneficial.}
    \label{fig:metatrain-analysis}
\end{figure}

To understand the method performance more clearly, we also add an ablation study where we compare the meta-test performance of policies at different iterations along meta-training. This shows the effect that additional meta-training has on the fast learning performance for new tasks. This comparison is shown in Figure~\ref{fig:metatrain-analysis}. As can be seen here, at iteration 0 of meta-training the policy is not a very good initialization for learning new tasks. As we move further along the meta-training process, we see that the meta-learned initialization becomes more and more effective at learning new tasks. This shows a clear correlation between additional meta-training and improved meta test-time performance. 

\subsection{Analysis of Learned Task Distributions}
\label{sec:task-analysis}

\begin{figure*}[t]
    \centering
        \begin{subfigure}[b]{0.3\textwidth}
          \centering\includegraphics[width=\textwidth, height=0.7\textwidth]{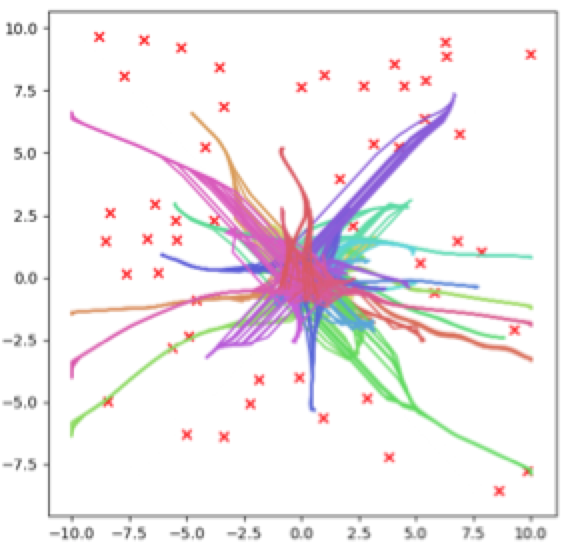}
          \caption*{2D navigation}
        \end{subfigure}
        \begin{subfigure}[b]{0.3\textwidth}
          \centering\includegraphics[width=\textwidth]{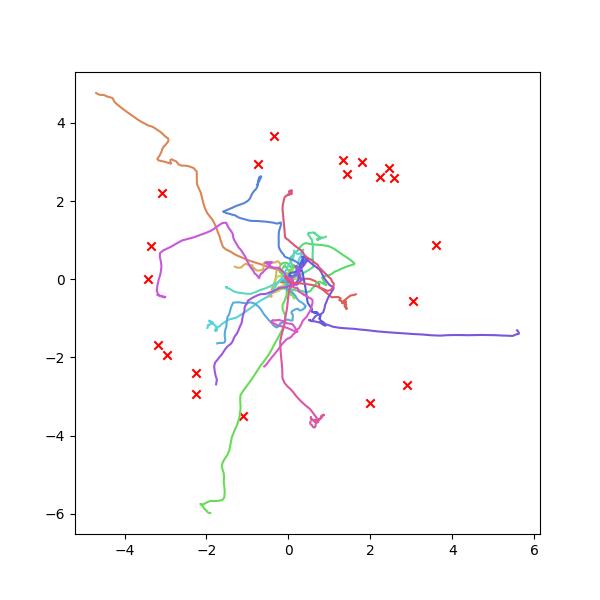}
          \caption*{Ant}
        \end{subfigure}
        \begin{subfigure}[b]{0.3\textwidth}
          \centering\includegraphics[width=\textwidth, height=0.75\textwidth]{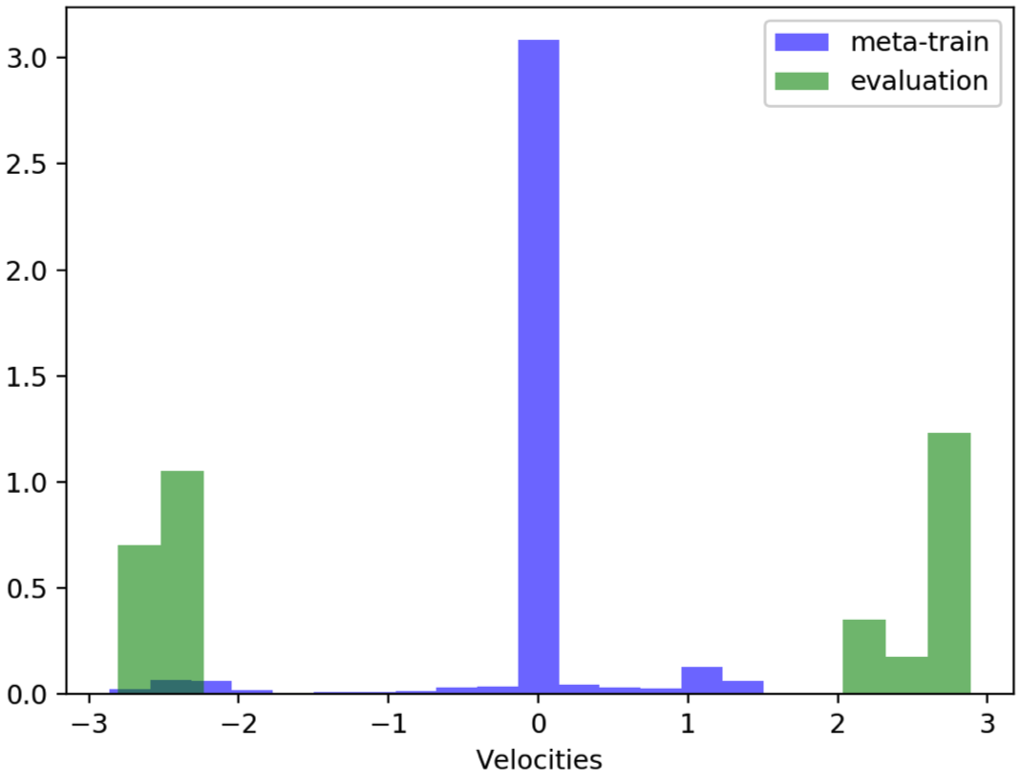}
          \caption*{Half-Cheetah}
        \end{subfigure}
        \vspace{-0.2cm}
    \caption{\footnotesize \textbf{Learned meta-training task distribution and evaluation tasks}: We plot the center of mass for various skills discovered by point mass and ant using DIAYN, and a blue histogram of goal velocities for cheetah. Evaluation tasks, which are not provided to the algorithm during meta-training, are plotted as red `x' for ant and pointmass, and as a green histogram for cheetah. While the meta-training distribution is broad, it does not fully cover the evaluation tasks. Nonetheless, meta-learning on this \emph{learned} task distribution enables efficient learning on a test task distribution.}
    \label{fig:train-distribution}
\end{figure*}

We can analyze the tasks discovered through unsupervised exploration and compare them to tasks we evaluate on at meta-test time. Figure~\ref{fig:train-distribution} illustrates these distributions using scatter plots for 2D navigation and the Ant, and a histogram for the HalfCheetah. Note that we visualize dimensions of the state that are relevant for the evaluation tasks -- positions and velocities -- but these dimensions are \emph{not} specified in any way during unsupervised task acquisition, which operates on the entire state space. Although the tasks proposed via unsupervised exploration provide fairly broad coverage, they are clearly quite distinct from the meta-test tasks, suggesting the approach can tolerate considerable distributional shift. Qualitatively, many of the tasks proposed via unsupervised exploration such as jumping and falling that are not relevant for the evaluation tasks. Our choice of the evaluation tasks was largely based on prior work, and therefore not tailored to this exploration procedure. The results for unsupervised meta-RL therefore suggest quite strongly that unsupervised task acquisition can provide an effective meta-training set, at least for MAML, even when evaluating on tasks that do not closely match the discovered task distribution.

\section{Hyperparameter Details}

\begin{figure}[!h]
    \centering
        \begin{subfigure}[b]{0.48\linewidth}
          \centering\includegraphics[width=\linewidth]{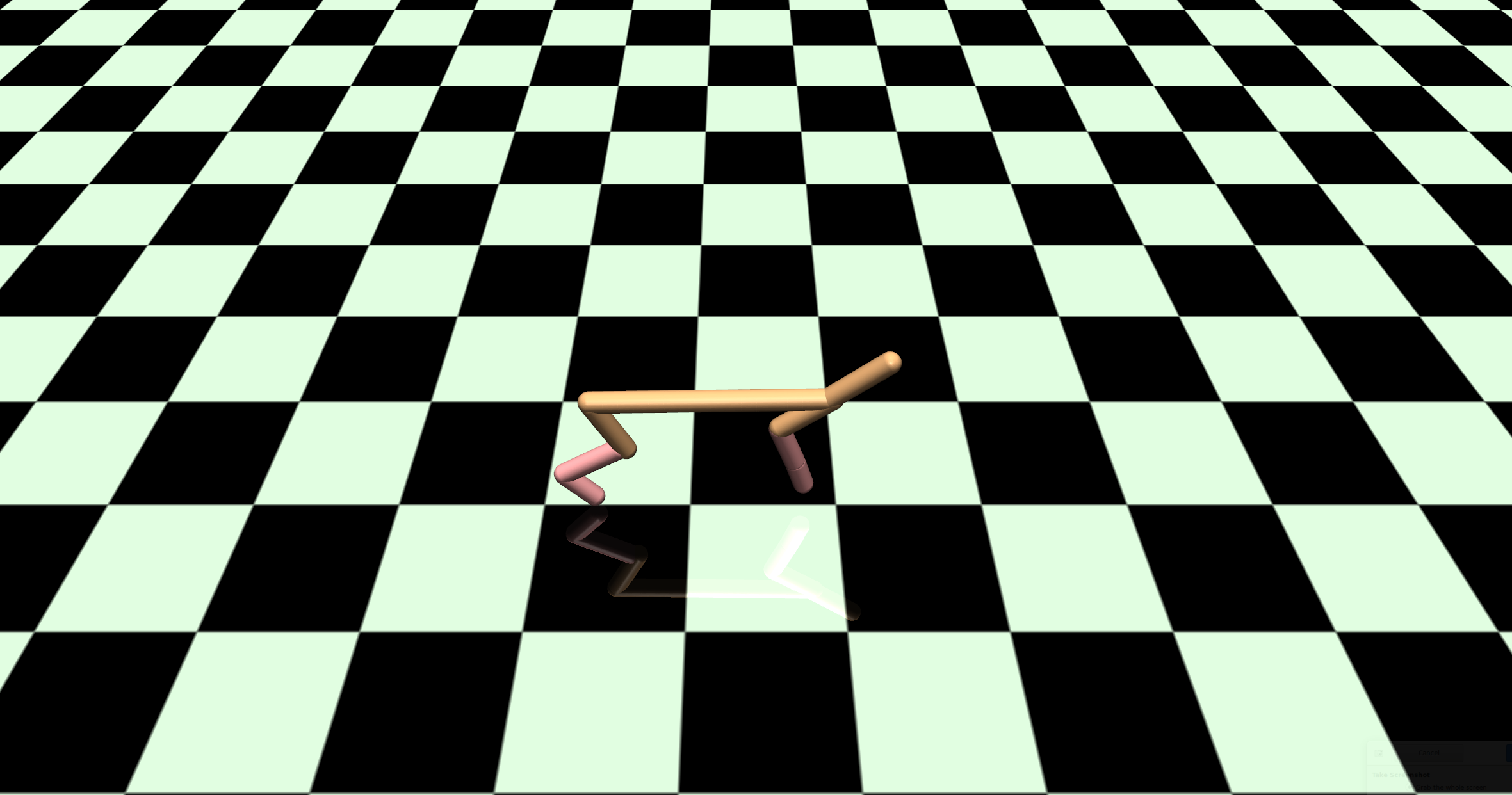}
          \caption*{Half-Cheetah}
        \end{subfigure}%
        ~
        \begin{subfigure}[b]{0.48\linewidth}
          \centering\includegraphics[width=\linewidth]{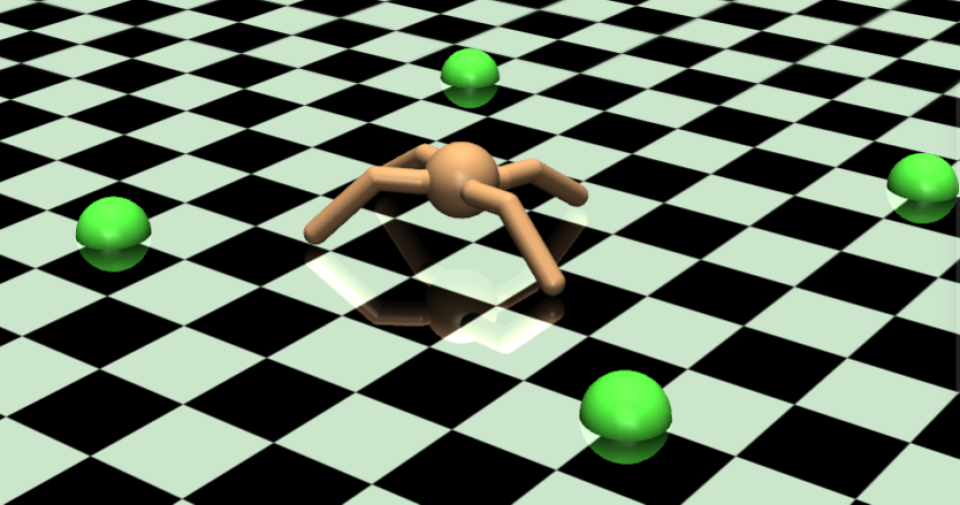}
          \caption*{Ant}
       \end{subfigure}
      \caption{\textbf{Environments}: \figleft \; Half-Cheetah and \figright \; Ant} \label{fig:envs}  
\end{figure}

For all our experiments, we used DIAYN to acquire the task proposals using $20$ skills for half-cheetah and for ant and $50$ skills for the 2D navigation. We illustrate these half cheetah and ant in Fig.~\ref{fig:envs}. We ran the domains using the standard DIAYN hyperparameters described in \url{https://github.com/ben-eysenbach/sac} to acquire task proposals. These proposals were then fed into the MAML algorithm \url{https://github.com/cbfinn/maml_rl}, with inner learning rate $0.1$, meta learning rate $0.01$, inner batch size $40$, outer batch size, path length $100$, using 2 layer networks with 300 units each with ReLu nonlinearities. We vary the meta-batch size according to the number of skills: 50 for pointmass, 20 for cheetah, and 20 ant. The test time learning is done with the same parameters for the UMRL variants, and done using REINFORCE with the Adam optimizer for the comparison with learning from scratch. We swept over learning rates for learning from scratch via vanilla policy gradient, and found that using ADAM with adaptive step size is the most stable and quick at learning. 

\clearpage

\end{document}